\documentclass[letterpaper, 10 pt, conference]{ieeeconf}

\IEEEoverridecommandlockouts                              

\usepackage{graphicx} 
\usepackage{amsmath}
\usepackage{amssymb}
\usepackage{booktabs}
\usepackage[table]{xcolor} 
\usepackage{siunitx} 
\usepackage[para,online,flushleft]{threeparttable}
\usepackage{multirow}
\usepackage{url} 
\usepackage{svg}
\usepackage{hyperref}
\hypersetup{
    colorlinks=true,
    linkcolor=blue,
    filecolor=magenta,      
    urlcolor=cyan,
    pdftitle={Overleaf Example},
    pdfpagemode=FullScreen,
    }
\usepackage{tikz}
\usetikzlibrary{matrix,calc}
\title{\LARGE \bf

FunGraph: Functionality Aware 3D Scene Graphs 
\\ for Language-Prompted Scene Interaction
}

\author{Dennis Rotondi, Fabio Scaparro, Hermann Blum, Kai O. Arras
\thanks{D. Rotondi, F. Scaparro and K.O. Arras are with the Socially Intelligent Robotics Lab, Institute for Artificial Intelligence, University of Stuttgart, Germany. Email: {\tt\small dennis.rotondi@ki.uni-stuttgart.de}. \indent H. Blum is with the Robot Perception and Learning Lab, LAMARR Institute for Machine Learning and Artificial Intelligence, University of Bonn, Germany.}%
\thanks{This work is partially supported by the Robotics Institute Germany (RIG) funded by BMBF. The authors also thank the International Max Planck Research School for Intelligent Systems (IMPRS-IS) for supporting Dennis Rotondi. }
}

\date{}
\newcommand{\website}[0]{\href{https://fungraph.github.io}{https://fungraph.github.io}}
\newcommand{\acceptance}{%
\begin{tikzpicture}[overlay, remember picture]
\path (current page.north) ++(0,-1cm) node[align=center] {
This paper has been accepted for publication at the 2025 International Conference on Intelligent Robots and Systems (IROS).
};
\end{tikzpicture}
}
\newcommand{\isArxiv}[2]{#1} %

\begin{document}




\newcommand{\vect}[1]{\boldsymbol{#1}}
\newcommand{\mat}[1]{\boldsymbol{#1}}

\newcommand{\diff}[2]{\frac{\partial #1}{\partial #2}}
\newcommand{\diffs}[3]{\frac{\partial^2 #1}{
\ifx#2#3 
\partial #2^2
\else
\partial #2 \partial #3
\fi
}}
\newcommand{\de}[1]{{\left | {#1} \right |}}

\newcommand{\dddott}[1]{{\stackrel{\boldsymbol{...}}{#1}}}
\newcommand{\ddddott}[1]{{\stackrel{\boldsymbol{....}}{#1}}}

\newcommand{\normm}[1]{{\left \| {#1} \right \|}}


\newcommand{\zerov}{\vect{0}}

\newcommand{\av}{\vect{a}}
\newcommand{\bv}{\vect{b}}
\newcommand{\cv}{\vect{c}}
\newcommand{\dcv}{\dot{\vect{c}}}
\newcommand{\ddcv}{\ddot{\vect{c}}}
\newcommand{\dv}{\vect{d}}
\newcommand{\ev}{\vect{e}}
\newcommand{\dev}{\dot{\vect{e}}}
\newcommand{\ddev}{\ddot{\vect{e}}}
\newcommand{\fv}{\vect{f}}
\newcommand{\gv}{\vect{g}}
\newcommand{\gbv}{\bar{\vect{g}}}
\newcommand{\dgv}{\dot{\vect{g}}}
\newcommand{\ddgv}{\ddot{\vect{g}}}
\newcommand{\hv}{\vect{h}}
\newcommand{\kv}{\vect{k}}
\newcommand{\lv}{\vect{l}}
\newcommand{\mv}{\vect{m}}
\newcommand{\nv}{\vect{n}}
\newcommand{\dnv}{\dot{\vect{n}}}
\newcommand{\ddnv}{\ddot{\vect{n}}}
\newcommand{\ov}{\vect{o}}
\newcommand{\pv}{\vect{p}}
\newcommand{\dpv}{\dot{\vect{p}}}
\newcommand{\ddpv}{\ddot{\vect{p}}}
\newcommand{\qv}{{\vect{q}}}
\newcommand{\dqv}{\dot{\vect{q}}}
\newcommand{\ddqv}{\ddot{\vect{q}}}
\newcommand{\dddqv}{\dddott{\vect{q}}}
\newcommand{\ddddqv}{\ddddott{\vect{q}}}
\newcommand{\qbv}{\bar{\vect{q}}}
\newcommand{\dqbv}{\dot{\bar{\vect{q}}}}
\newcommand{\ddqbv}{\ddot{\bar{\vect{q}}}}
\newcommand{\dddqbv}{\dddott{\bar{\vect{q}}}}
\newcommand{\ddddqbv}{\ddddott{\bar{\vect{q}}}}
\newcommand{\qhv}{\hat{\vect{q}}}
\newcommand{\qtbv}{\tilde{\bar{\vect{q}}}}
\newcommand{\rv}{{\vect{r}}}
\newcommand{\drv}{\dot{\vect{r}}}
\newcommand{\sv}{\vect{s}}
\newcommand{\dsv}{\dot{\vect{s}}}
\newcommand{\uv}{\vect{u}}
\newcommand{\vv}{\vect{v}}
\newcommand{\dvv}{\dot{\vect{v}}}
\newcommand{\wv}{\vect{w}}
\newcommand{\dwv}{\dot{\vect{w}}}
\newcommand{\xv}{\vect{x}}
\newcommand{\dxv}{\dot{\vect{x}}}
\newcommand{\ddxv}{\ddot{\vect{x}}}
\newcommand{\dddxv}{\dddott{\vect{x}}}
\newcommand{\ddddxv}{\ddddott{\vect{x}}}
\newcommand{\txv}{\vect{\tilde{x}}}
\newcommand{\dtxv}{\dot{\tilde{\vect{x}}}}
\newcommand{\ddtxv}{\ddot{\tilde{\vect{x}}}}
\newcommand{\dddtxv}{\dddott{\tilde{\vect{x}}}}
\newcommand{\ddddtxv}{\ddddott{\tilde{\vect{x}}}}
\newcommand{\yv}{\vect{y}}
\newcommand{\ybv}{\bar{\vect{y}}}
\newcommand{\dyv}{\dot{\vect{y}}}
\newcommand{\ytv}{\tilde {\vect{y}}}
\newcommand{\zv}{\vect{z}}
\newcommand{\dzv}{\dot{\vect{z}}}
\newcommand{\ddzv}{\ddot{\vect{z}}}


\newcommand{\alphav}{\vect{\alpha}}
\newcommand{\betav}{\vect{\beta}}
\newcommand{\gammav}{\vect{\gamma}}
\newcommand{\lambdav}{\vect{\lambda}}
\newcommand{\muv}{\vect{\mu}}
\newcommand{\etav}{\vect{\eta}}
\newcommand{\deltav}{\vect{\delta}}
\newcommand{\ddeltav}{\dot{\vect{\delta}}}

\newcommand{\phiv}{\vect{\phi}}
\newcommand{\dphiv}{\dot{\vect{\phi}}}
\newcommand{\ddphiv}{\ddot{\vect{\phi}}}

\newcommand{\psiv}{\vect{\psi}}

\newcommand{\sigmav}{\vect{\sigma}}
\newcommand{\dsigmav}{\dot{\vect{\sigma}}}

\newcommand{\piv}{\vect{\pi}}

\newcommand{\tauv}{\vect{\tau}}
\newcommand{\dtauv}{\dot{\vect{\tau}}}
\newcommand{\ddtauv}{\ddot{\vect{\tau}}}
\newcommand{\thetav}{\vect{\theta}}
\newcommand{\dthetav}{\dot{\vect{\theta}}}
\newcommand{\ddthetav}{\ddot{\vect{\theta}}}
\newcommand{\thetavt}{\tilde{\vect{\theta}}}
\newcommand{\nuv}{\vect{\nu}}
\newcommand{\omegav}{\vect{\omega}}
\newcommand{\xiv}{\vect{\xi}}
\newcommand{\dxiv}{\dot{\vect{\xi}}}


\newcommand{\Fv}{\vect{F}}
\newcommand{\Mv}{\vect{M}}
\newcommand{\Tv}{\vect{T}}
\newcommand{\Vv}{\vect{V}}
\newcommand{\Wv}{\vect{W}}


\newcommand{\Alphav}{\vect{\Alpha}}
\newcommand{\Betav}{\vect{\Beta}}
\newcommand{\Gammav}{\vect{\Gamma}}
\newcommand{\Thetam}{\vect{\Theta}}
\newcommand{\dThetam}{\dot{\Thetam}}
\newcommand{\ddThetam}{\ddot{\Thetam}}


\newcommand{\IIm}{\mat{I}}
\newcommand{\zerom}{\mat{O}}

\newcommand{\Am}{\mat{A}}
\newcommand{\Bm}{\mat{B}}
\newcommand{\Cm}{\mat{C}}
\newcommand{\dCm}{\dot{\Cm}}
\newcommand{\ddCm}{\ddot{\Cm}}
\newcommand{\dddCm}{\dddott{\Cm}}
\newcommand{\Dm}{\mat{D}}
\newcommand{\Em}{\mat{E}}
\newcommand{\Fm}{\mat{F}}
\newcommand{\Gm}{\mat{G}}
\newcommand{\Hm}{\mat{H}}
\newcommand{\Jm}{\mat{J}}
\newcommand{\Lm}{\mat{L}}
\newcommand{\Jbm}{\bar{\mat{J}}}
\newcommand{\dJm}{\dot{\Jm}}
\newcommand{\ddJm}{\ddot{\Jm}}
\newcommand{\dddJm}{\dddott{\Jm}}
\newcommand{\Km}{\mat{K}}
\newcommand{\Mm}{\mat{M}}
\newcommand{\dMm}{\dot{\Mm}}
\newcommand{\ddMm}{\ddot{\Mm}}
\newcommand{\dddMm}{\dddott{\Mm}}
\newcommand{\Nm}{\mat{N}}
\newcommand{\Pm}{\mat{P}}
\newcommand{\Qm}{\mat{Q}}
\newcommand{\Rm}{\mat{R}}
\newcommand{\Sm}{\mat{S}}
\newcommand{\Tm}{\mat{T}}
\newcommand{\Vm}{\mat{V}}
\newcommand{\Um}{\mat{U}}
\newcommand{\Wm}{\mat{W}}
\newcommand{\Xm}{\mat{X}}
\newcommand{\Ym}{\mat{Y}}
\newcommand{\Zm}{\mat{Z}}


\newcommand{\Deltam}{\mat{\Delta}}
\newcommand{\Sigmam}{\mat{\Sigma}}
\newcommand{\Gammam}{\mat{\Gamma}}
\newcommand{\Lambdam}{\mat{\Lambda}}


\newcommand{\calAm}{\mat{\cal A}}
\newcommand{\calCm}{\mat{\cal C}}
\newcommand{\calMm}{\mat{\cal M}}
\newcommand{\dcalMm}{\dot{\calMm}}


\newcommand{\Jqbm}{\Jm_{\bar q}}
\newcommand{\Jgm}{\Jm_{g}}
\newcommand{\Jhm}{\Jm_{h}}



\newcommand{\calA}{{\cal A}}
\newcommand{\calB}{{\cal B}}
\newcommand{\calC}{{\cal C}}
\newcommand{\calE}{{\cal E}}
\newcommand{\calF}{{\cal F}}
\newcommand{\calG}{{\cal G}}
\newcommand{\calI}{{\cal I}}
\newcommand{\calM}{{\cal M}}
\newcommand{\calO}{{\cal O}}
\newcommand{\calT}{{\cal T}}
\newcommand{\calV}{{\cal V}}

\newcommand{\eg}{\emph{e.g.,}\xspace}
\newcommand{\ie}{\emph{i.e.,}\xspace}

\newcommand{\PCD}{\emph{PCD}}

\maketitle
\isArxiv{\acceptance}{}
\thispagestyle{empty}
\pagestyle{empty}

\begin{abstract}
The concept of 3D scene graphs is increasingly recognized as a powerful semantic and hierarchical representation of the environment. Current approaches often address this at a coarse, object-level resolution.
In contrast, our goal is to develop a representation that enables robots to directly interact with their environment by identifying both the location of functional interactive elements and how these can be used.
To achieve this, we focus on detecting and storing objects at a finer resolution, focusing on affordance-relevant parts.
The primary challenge lies in the scarcity of data that extends beyond instance-level detection and the inherent difficulty of capturing detailed object features using robotic sensors.
We leverage currently available 3D resources to generate 2D data and train a detector, which is then used to augment the standard 3D scene graph generation pipeline. 
Through our experiments, we demonstrate that our approach achieves functional element segmentation comparable to state-of-the-art 3D models and that our augmentation enables task-driven affordance grounding with higher accuracy than the current solutions.
See our project page at \website.
\end{abstract}

\section{Introduction}
A semantically rich 3D representation of the environment is essential for various robotic downstream tasks, including task-driven perception, planning, or manipulation. Building on the capability of SLAM to learn topometric maps and leveraging recent advancements in multimodal foundation models, the concept of 3D scene graphs is gaining increasing attention. 3D scene graphs (3DSG) \cite{armeni20193dscenegraphstructure, Kim_2020} are hierarchical representations that capture both the geometric and semantic structure of a scene, where nodes represent objects or spaces at various scales, and edges define their relationships. They enable robots to understand and reason about their surroundings, achieving unprecedented levels of open-vocabulary 3D scene understanding \cite{conceptgraph,hov-sg}.

In related work, 3D scene graphs are typically structured at the room and object level (e.g., \cite{conceptgraph}). However, this level of granularity may be insufficient for robots that physically interact with their environment. For typical tasks in a household such as opening a fridge, adjusting a thermostat, or operating switches and buttons, a more fine-grained representation is key for context-aware interactions. This should include objects and functional object parts such as knobs, handles or buttons, as well as predictions of possible interactions with them. In other words, in addition to inter-object relationships, the 3D scene graph should be extended by intra-object relationships, a problem largely underexplored in robotics and the focus of this paper, see Fig.~\ref{fig:ourmethodexample}.

One of the key challenges in modeling intra-object relationships is accurately perceiving functional object parts. Such parts are typically small, sparsely sampled in sensory data, and underrepresented in public datasets. As a result, even state-of-the-art object detectors struggle with low detection accuracy in both 2D and 3D. 

To address this challenge, we utilize SceneFun3D \cite{delitzas2024scenefun3d}, a large-scale dataset that provides sensory data and  3D annotations for functional interactive elements in household environments. 
However, this dataset was collected using expensive, highly accurate 3D LiDAR sensors, whereas our focus is on robotics scenarios that rely on affordable, on-board RGB-D sensors.
Therefore, our approach involves generating 2D data for detector fine-tuning and analyzing its impact in 3D. We leverage this result to extend 3D scene graphs with intra-object relationships, enabling task-driven affordance grounding.
Specifically, our contributions are:

\begin{figure}[t!]
  \centering
  \includegraphics[width=1.0\columnwidth,trim={0 0 0 0},clip]{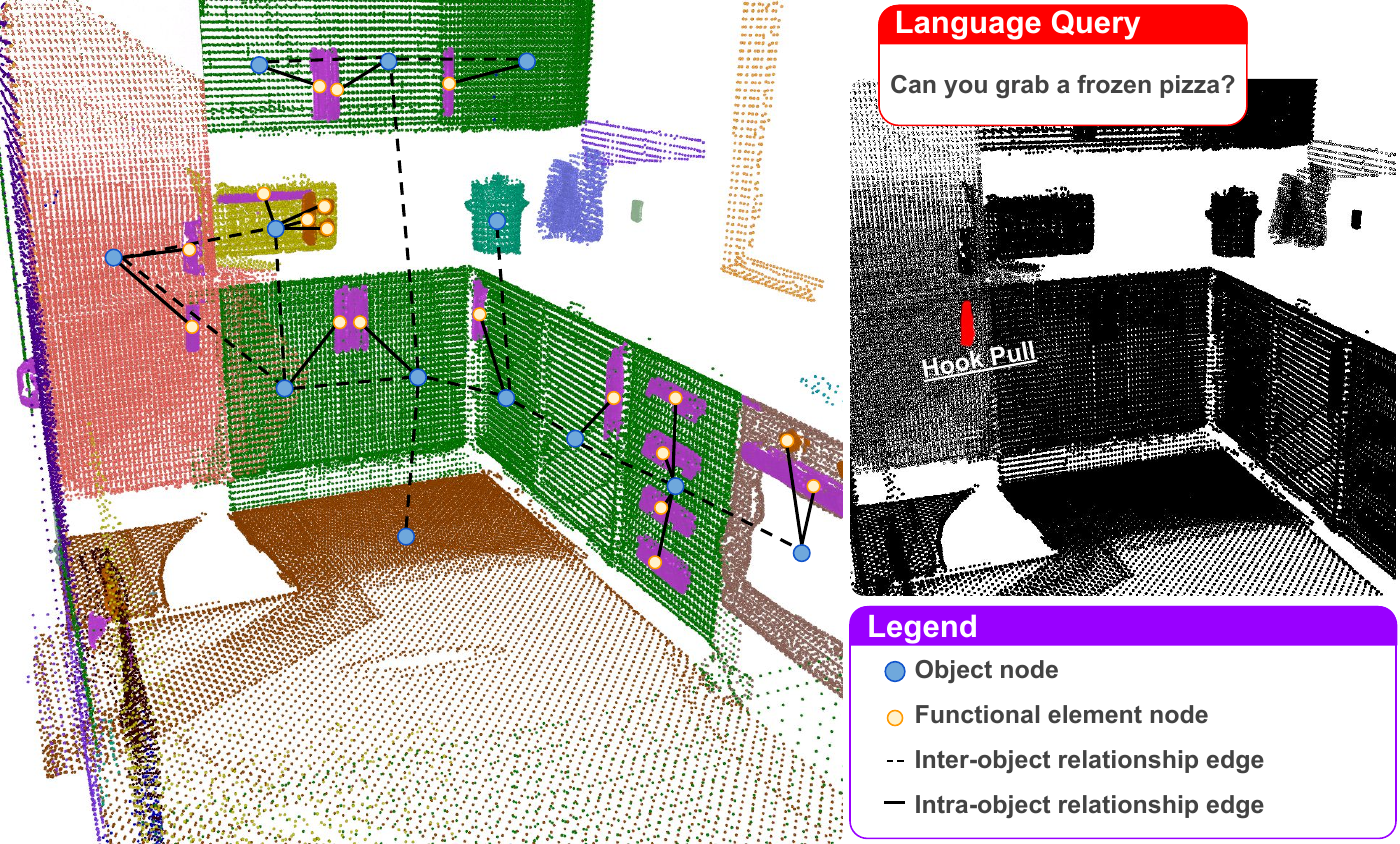}
   \caption{An example of a generated 3D scene graph and its application. The model represents both object and functional element nodes linked through intra-object relationships. All purple-colored segments in the left image have been correctly detected as functional elements of type ``Hook Pull,'' associated with different kitchen objects. One use of this representation is affordance grounding, in which unconstrained language queries produce the relation to the functional elements required to fulfill a task (freezer handle highlighted in red, right image).}
  \label{fig:ourmethodexample}
\end{figure}

\begin{itemize}
    \item A method to detect functional interactive elements from images, predict their affordances, and assign contextualized descriptions.
    \item A framework, FunGraph, for extending the generation of 3D scene graphs through intra-object relationships, representing functional elements and their affordances in relation to object nodes.
    \item A quantitative and qualitative evaluation of how the produced structure allows for 3D functional elements segmentation and task-driven affordance grounding. 
\end{itemize}

The source code and additional materials for FunGraph are publicly available at \website.

\section{Related Work}
\subsection{3DSG Generation and Prediction}\label{sec:rla}
3D scene graphs (3DSG)~\cite{armeni20193dscenegraphstructure, Kim_2020,rosinol20203ddynamicscenegraphs} are a spatial data representation in the form of a hierarchical graph with nodes representing different parts of a scene, such as buildings, rooms, and at the most granular level objects. These objects can then be connected through various relationships, including spatial (e.g., A ``is next to" B), comparative (e.g., A ``is larger than" B), and support (e.g., A ``lays on" B).

To build 3DSG from observations, different lines of work investigated the incremental creation from raw sensor inputs~\cite{conceptgraph, hov-sg, hydra, clio}, or create scene graphs from existing 3D scans in post-processing~\cite{armeni20193dscenegraphstructure, wald2020learning3dsemanticscene, wu2021scenegraphfusionincremental3dscene, koch2023sgrec3dselfsupervised3dscene, wang2023vlsatvisuallinguisticsemanticsassisted, koch2024open3dsgopenvocabulary3dscene}. The latter is sometimes referred to as 3DSG prediction. All these prior works have in common that their lowest layer is at the level of objects, and they do not consider object parts or their functionality and articulation.

Among the related works, only CLIO \cite{clio} mentions a need for different granularity, e.g., to represent that a piano is composed of keys and other parts. However, they only consider object parts and do not group them into objects. 
They further rely on point-grid-prompted SAM \cite{sam} to propose object segments, which is not reliably capable of detecting small interactive elements such as knobs. Another recent work~\cite{10897827} uses a 3DSG with object-to-object-part relations, but only in a single demo, to retrieve an object from a drawer.

\subsection{Alternative Queryable Scene Representations}
Implicit representations such as NeRFs \cite{nerf} are known for their rendering accuracy and, if not based on feature grids, have no explicitly defined resolution, which is ideal for functional interactive elements that are generally small. However, semantic variants of NeRFs \cite{Siddiqui_2023_CVPR, Zhi:etal:ICCV2021, lerf2023, engelmann2024opennerf} and Gaussian Splatting \cite{langsplat2023} are evidently limited with regard to small objects and sub-parts. These methods rely on pixel-aligned semantic features such as DINO~\cite{dino} or CLIP~\cite{clip}, which, however, lack the necessary level of detail for functional elements.

Search3D \cite{takmaz2025search3d} aims to build a hierarchical, open-vocabulary 3D scene representation that addresses finer-grained scene entities, such as object parts. 
However, its segmentation is purely geometric. Functional elements are small and, therefore, require high-resolution scans to be geometrically detectable, which has limited practicality.
Additionally, the method lacks semantic richness, which is heavily dependent on the foundation model employed. It does not establish any inter-relationships between entities in the scene, making queries such as, ``turn off the light \textit{above} the stove" impossible to answer.

\subsection{Functionality Segmentation \& Affordance Understanding}\label{sec:rlc}
The PartNet dataset \cite{Mo_2019_CVPR} consists of dense annotations on 3D CAD models, and datasets like 3D AffordanceNet \cite{9577339} and PartAfford \cite{xu2022partaffordpartlevelaffordancediscovery} have been built upon it, focusing on Gibsonian affordances \cite{gibson}, which characterize how humans interact with human-made objects and environments. These datasets annotate, for example, how a chair can be divided into backrest, seat, and legs. Prior works have linked this to robotic interaction~\cite{morlans2024grasp,schiavi2023learning}. 
Their focus, however, is primarily on predicting Gibsonian affordances of already isolated objects, rather than on our investigated problem of identifying functional interactive elements in a larger scene. 
MultiScan~\cite{mao2022multiscan} takes a step toward highlighting movable object parts in a room scan, such as individual drawers in a cabinet. However, it does not provide comprehensive annotations for interactive components like knobs, handles, and other similar elements in the scene.
The recent SceneFun3D \cite{delitzas2024scenefun3d} was the first dataset to annotate the functional elements themselves in real room-scale scenes. 
Based on ARKitScenes \cite{dehghan2021arkitscenes}, they select nine Gibsonian-inspired affordance labels to represent interactions with common elements in indoor environments (e.g., \textit{Rotate}, \textit{Hook Pull}, etc.) and annotate the 3D point cloud directly. Notably, this data annotates functional interactive elements in isolation, i.e., it lacks information on which knob opens which drawer or whether, e.g., two knobs would open the same drawer.
To the best of our knowledge, no 2D dataset with annotated functionally interactive elements currently exist. Therefore, prior 2D detectors, for example, leveraged semi-supervised learning from human video demonstrations~\cite{bahl2023affordances} to predict interactive elements.

\begin{figure}[t!]
  \centering
  \includegraphics[width=1\columnwidth,trim={0 0 0 0},clip]{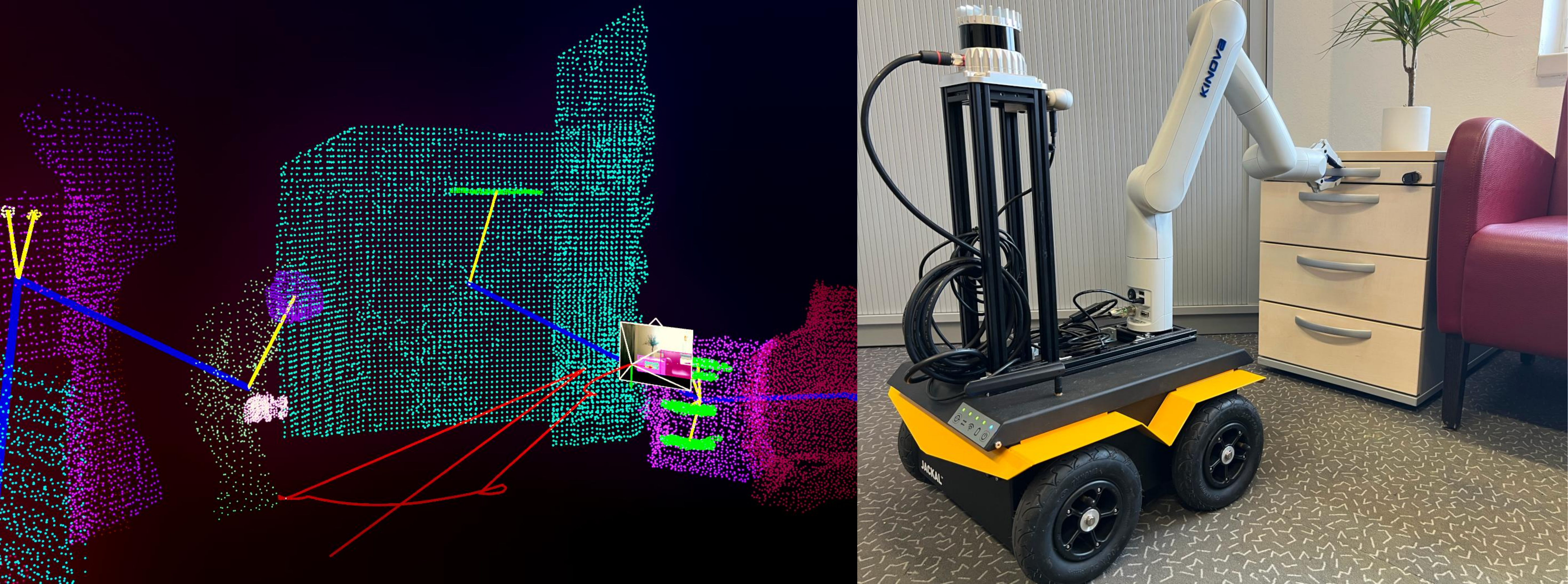}
  \caption{
  Implementation of our approach on a mobile manipulator using RGB-D sensing (3D LiDAR not used). The yellow edges in the left picture show intra-object relationships, the blue ones represent inter-object relationships, and the red curve is the robot's camera trajectory. On the right, our robot is shown grasping the top handle of the detected drawer cabinet.}
  \label{fig:realrobot}
\end{figure}

While predictors trained on large datasets offer important priors, it is actually impossible to know how, for example, a kitchen cabinet door opens without trying. Therefore, researchers have also looked into robotic
interactive exploration~\cite{buchanan2024online,ning2023where2explore} as a way to complement predictive priors. These works are directly compatible with the scene graph framework that we propose, where each interactive element prediction can be updated as a robot collects experience in an environment.
In the same way, our structure supports all solutions that learn a policy to interact with objects \cite{wang2025adamanip, donat2025fusingpointcloudvisual, zhu2024point} because they can be used as a final step to effectively perform the task (see Fig~\ref{fig:realrobot}).

\section{Method}

\subsection{Problem Formulation}
This work aims to extend the classical pipeline of 3D scene graph generation for indoor environments by incorporating intra-object relationships between scene objects (referred to as ``objects" in the following) and their 3D functionally interactive elements (referred to as ``functional elements" in the following). For example, we want a cabinet to have a direct relationship with its knobs, enhancing the scene graph with information about the object's possible interactions.

The input to the proposed method consists of a series of RGB-D observations, $\mathcal{I} = \{I_1, I_2, \dots, I_N\}$, and corresponding camera poses, $\mathcal{P}= \{\Pm_1, \Pm_2, \dots, \Pm_N\}$. Note that, in the following, each image \( I_i \) is captured from pose \( \Pm_i \) using camera parameters \( \Km_i \).

The output of the proposed method is a 3D scene graph, specifically a hierarchical graph $\calG = (\calV, \calE)$. That is, the set of nodes $\calV$ can be partitioned into $l$ layers ($\calV = \cup_{i=1}^{l} \calV_i$). Indeed, by design, introducing intra-object relationships does not interfere with the hierarchical properties of the structure: \textit{single parent}, \textit{locality}, and \textit{disjoint children} \cite{hydraextended}, which involve nodes $\calV$ (the collection of objects of the 3D scene) and edges $\calE$ (the relationships between objects) of the graph.

To ensure the validity of all the theoretical results from Hughes et al. \cite{hydraextended}, care must be taken with the property of disjoint children. Assuming that the objects are relatively small within the scene, any spatial relationship $s_e \in \mathcal{E}$ can be directly inherited by these smaller objects in the 3D scene graph representation without losing any information about the scene. 
For example, if a wardrobe is next to a closet, we can also infer that the knobs of the closet are next to the wardrobe without the need to add explicit edges between the wardrobe and the functional elements of the closet. However, this does not hold for comparative relationships. For instance, ensuring the disjoint children property prevents us from relating functional elements of two different objects.  

In this work, we consider only spatial relationships for our graph. That said, it should be noted that incorporating non-spatial relationships (e.g., between a light switch and a light) may break the hierarchical graph structure.
\subsection{Generation of 2D Data}\label{sec:methb}
By definition, functional elements are meant for human interaction. Therefore, common functional interactive elements such as door knobs, light switches, or handles are all smaller than a regular human hand. To detect these elements, robots require sensors with sufficient spatial resolution.

Because cm-resolution LiDARs are not detailed enough and mm-resolution 3D scanners are cost-prohibitive in many robotic applications, we assume a collection of registered RGB-D observations as input and propose to detect functional elements from high-resolution RGB.

To detect functional elements from RGB images, we find state-of-the-art open-vocabulary object detectors \cite{yolo, ren2024groundingdino15advance} to be insufficient (cf. Table~\ref{tab:detection2d}).
Since there are also no other pre-trained detectors for this task, our first step is to produce data for our own model. In particular, we preprocess the datasets listed in \ref{sec:rlc} to be used as 2D training data.

The procedure illustrated in this section is applied to SceneFun3D \cite{delitzas2024scenefun3d}, as it is the only resource at the time of writing containing detailed information about functional elements. In any case, this approach is general and can be applied to similar data in the future to generate robust 2D observations from annotated 3D point clouds.

For each scene point cloud \( \PCD \) in the dataset, along with 3D functional interactive element annotations \( \mathcal{A} = \{\Am_1, \Am_2, \dots, \Am_M\} \), where each annotation represents a single point cloud segment \( \Am_j \subset \PCD \), and RGB-D observations \( \mathcal{I} \) captured from the scene, the goal is to generate 2D annotated images with bounding boxes for performing 2D object detection of functional interactive elements.

For each image $I_i$, each annotation \( \Am_j \) is projected on the 2D image plane as $\av_j = (\xv_j \; \yv_j \; \zv_j )^T = \Km_i \Pm_i^{-1} \Am_j$
Then, we mask out all points from \( \av_j \) that are behind (\( z_{j} < 0 \)), or outside of the current camera image.
We also remove all points where the depth \( \dv_i \) in \( I_i \) differs from the projected depth \( \zv_j \) by more than a certain threshold \( \theta_{\emph{depth}}\), indicating that this annotated element is occluded in the current frame. Finally, if the annotation projected onto the image has a bounding box area larger than a certain threshold \( \theta_{\emph{area}}\) and the ratio of pixel points to total points in \( \Am_j \) exceeds a threshold \( \theta_{\emph{points}}\), it is kept; otherwise, it is discarded.
At the end of the process, each image \( I_i \) associated with a scene will be left with a subset of projected annotations from \( \mathcal{A}^{[i]} \subset \mathcal{A} \), whose bounding boxes will contribute to the 2D dataset (the collection of all annotated images obtained through this process from all scene point clouds) used to train the 2D detector for functional elements.

\subsection{Functionality Aware 3D Scene Graph Generation}\label{sec:methc}
\begin{figure*}[th!]
    \centering
    \includegraphics[width=1.0\textwidth,trim={0.35cm 0 0.3cm 0},clip]{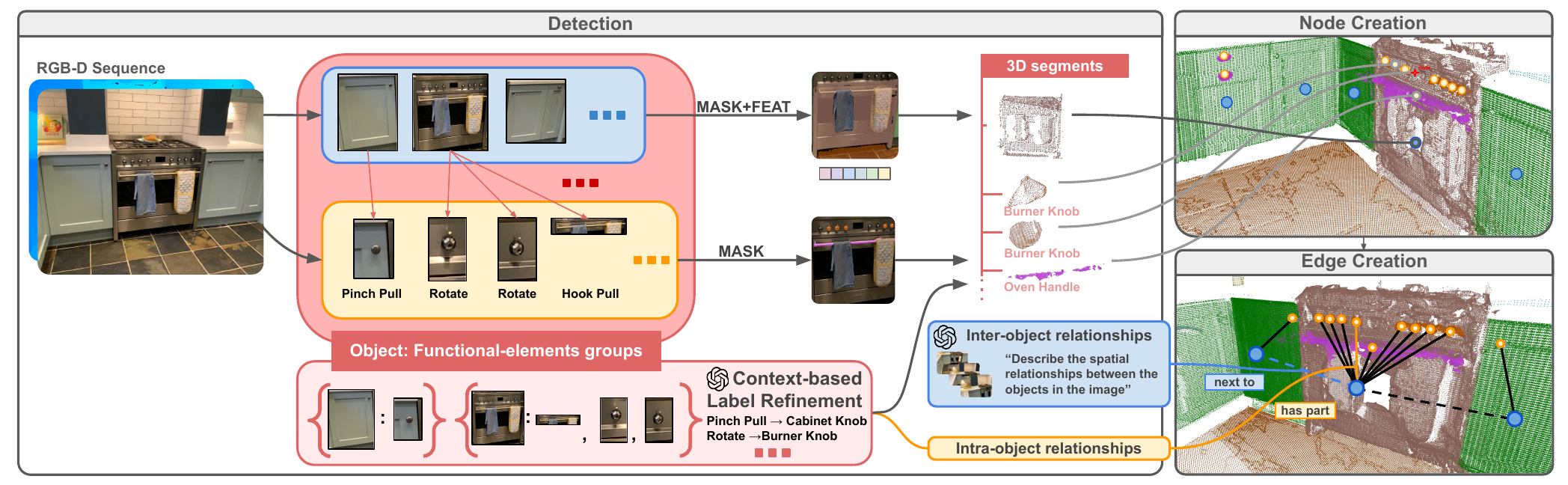}
    \caption{Overview of our functionality-aware 3D scene graph generation pipeline, which consists of three stages: (1) Detection, where instance segmentation and feature extraction are performed to identify both objects and functional elements; (2) Node Creation, where multi-view geometric and semantic feature merging is used to construct graph nodes; and (3) Edge Creation, where inter-object and intra-object relationships are established using spatial reasoning and affordance-based associations. }\label{fig:methodoverview} 
\end{figure*}

In the past few years, a clear methodology has emerged as a framework for generating a 3D scene graph from RGB-D observations \cite{conceptgraph, hov-sg, hydra, kassab2024barenecessitiesdesigningsimple}. The process consists of three phases: 
\begin{itemize}
    \item {\bf Detection:} Instance segmentation of entities captured in the images and feature extraction.
    \item {\bf Node creation:} Multi-view geometric and semantic feature merging.
    \item {\bf Edge creation:} Relationship generation.
\end{itemize}
Each of these phases requires particular attention when dealing with objects the size of functional elements.
Fig.~\ref{fig:methodoverview} provides an overview of our method.

\textbf{Detection.}
Rather than the class-agnostic 2D segmentation of \cite{hov-sg}, we opt for the approach from ConceptGraphs \cite{conceptgraph}. 
This method detects objects before segmenting them for each image $I_i$, making it straightforward to associate objects with their functional elements. The class-agnostic segmentation was shown to produce slightly better results in object retrieval, but without clearly defined instances, the association between objects and functional elements is more complicated and potentially less reliable.

For each \( I_i \in \mathcal{I} \), the \( q \) classes and bounding boxes of objects and functional elements are detected independently using two different models.
 
After filtering out detections with confidence below $\theta_{\text{bbox}}$ and relative-to-image area ratio lower than $\theta_{\text{rarea}}$, we prompt a general-purpose segmentation model with each remaining bounding box. The segments are then reprojected into 3D using the depth information of $I_i$ and denoised using DBSCAN clustering, resulting in a set of 3D object segments (or point clouds)
\[
\calO^{[i]} = \{\calO^{[i]}_m, \calO^{[i]}_a\} = \{\ov^{[i]}_{m1},\dots, \ov^{[i]}_{mp},\ov^{[i]}_{ap+1},\dots, \ov^{[i]}_{aq}\}
\] 
where the subscript $m$ denotes objects and $a$ denotes functional elements. Each object is associated with a label $c^{[i]}_j$ and semantic features $\fv^{[i]}_{\!j}$ extracted using the CLIP model from the bounding-box cropped image. We further discard functional elements that do not overlap with any object's bounding box by at least $\theta_{\text{rel}}$.

\begin{figure}[tb!]
  \centering
  \includegraphics[scale=0.4, trim={2.2cm 0 2.3cm 1cm},clip]{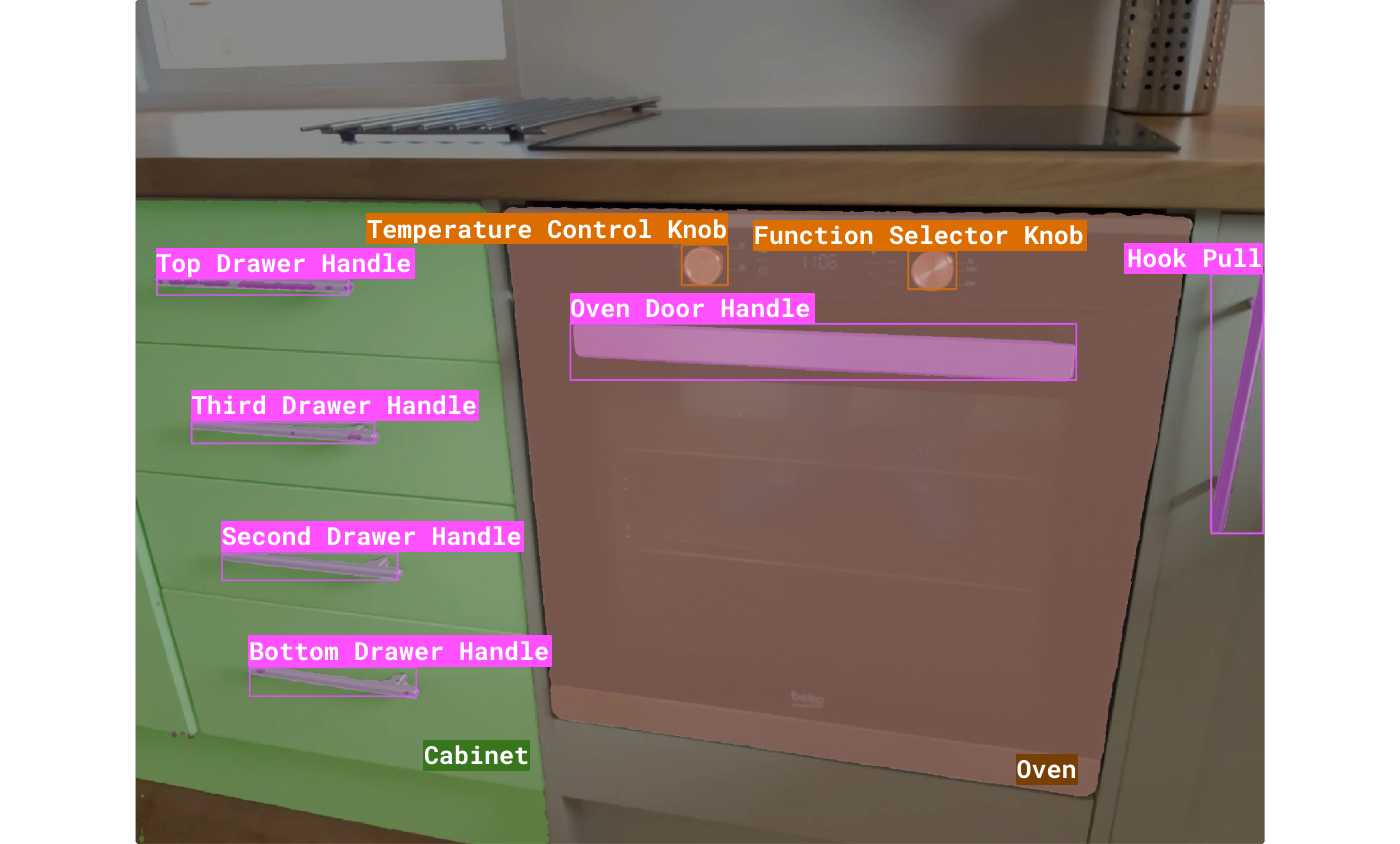}
  \caption{Illustration of our context-based label refinement. The VLM is queried to contextualize functional elements with their associated parent objects. The handle on the right, labeled ``Hook Pull," is not linked to any object, and, as a result, cannot be refined further.
  }
  \label{fig:lablerefinement}
\end{figure}

\textit{Context-Based Label Refinement.} With the detector trained using the resource in \ref{sec:methb}, functional elements are classified into a closed set of Gibsonian-inspired affordances based on the SceneFun3D annotations. To ensure the method’s open vocabulary ability and obtain more concrete descriptions of the functional elements, we prompt a visual language model (VLM) for each group of object and its associated functional elements. 
For example, this allows us to obtain the names ``Refrigerator Handles" and ``Freezer Handle" from an image of a refrigerator, which is associated with two ``Hook Pull" functional elements through bounding box overlap.
Each element in the image prompt is annotated using the bounding box and the current label name. Fig.~\ref{fig:lablerefinement} shows an example result.

\textbf{Node Creation.} In the following, we incrementally merge the object detections from images to create the graph's nodes.
For each image, each $\ov^{[i]}_{mj} \in \calO^{[i]}_m$ is compared to the nodes' point cloud in the graph representing objects. 
To be merged (though union), two point clouds $u,v$ need to score over $\theta_\emph{geo}$ according to a geometric similarity score and over $\theta_\emph{sem}$ according to a semantic similarity score. The geometric similarity is \cite{hov-sg}: 
\begin{align}\label{eq:geosim}
S(u,v)=\max(\emph{overlap}(u,v), \emph{overlap}(v,u))
\end{align}
where $\emph{overlap}(u,v)$ is the ratio of the points of $u$ withing a certain distance from $v$ divided by the total number of points of $u$.
For the semantic comparison, cosine similarity $\phi(\fv_n, \fv^{[i]}_{j})$ between the features of the node $n$ and the object is used. 
The object segment is merged into the node that yields the highest sum of the two similarities in the described process.
After each successful merge, the point cloud of node $n$ is denoised using DBSCAN and downsampled to reduce redundancy, and then the semantic features are updated as 
\begin{align}
\fv_n \leftarrow (\fv_n*n_n + \fv^{[i]}_{j})/(n_n+1)
\end{align}
where \( n_n \) represents the number of times a point cloud has been merged into the one of node \( n \).  
The class \( c^{[i]}_j \) is saved along with all the other classes merged into node \( n \), and only after processing all the images is the most common class assigned.
If no similarity check is passed, \( \ov^{[i]}_{mj} \) is merged with an empty, newly initialized node containing a zero-feature vector and an empty associated point cloud.
Then, for each \( \ov^{[i]}_{ak} \) associated with \( \ov^{[i]}_{mj} \), which was previously merged into node \( n \), we search for the most suitable node among the associated functional element nodes of the 3DSG related to \( n \) to merge it into.  
Only the metric \eqref{eq:geosim} is employed, but with a different threshold \( \theta_{\emph{geo2}} \). The same merging process is followed, with two key differences: redundant points are removed without downsampling, and the functional element segment is merged with all nodes that pass the similarity check. This approach is beneficial because, when dealing with objects the size of functional elements, they are often only partially observed, making it difficult to fill certain gaps that could otherwise lead to a final merge at a later stage.

If no similarity check is passed, the functional element is merged into a newly initialized empty node and associated with node \( n \).

Periodically, the graph nodes are processed in batches and merged with each other, first merging object nodes and then functional element nodes related to the merged object nodes if they pass the geometric and semantic similarity checks.

When all images and their corresponding elements have been processed, we remove nodes that are not formed by at least \( \theta_{\text{num}} \) merged segments from the graph. 
For the remaining functional element nodes, we apply an outlier removal procedure using radial and statistical filters. This enhances the robustness of the graph creation process by preventing noisy reconstructions from being included.

\textbf{Edge Creation.} Our 3DSG has two types of edges: spatial inter-object relationships between the objects in the scene and intra-object relationships between objects and their functional elements.

For \textit{inter-object relationships}, during the node creation phase, we track the image sources of all object segments that contribute (through merging) to different nodes. We then sequentially prompt a VLM with images where all detected objects are highlighted to extract binary relationships between objects \( (i, j) \).

The most common relationship is added as an edge between the nodes of \( i \) and \( j \), encoding this spatial knowledge. 
The relationship can be either symmetric (e.g., ``next to'') or directional (e.g., ``on top of'').

For \textit{intra-object relationships}, we use the bounding box association described in the detection phase to establish a ``has-part" directed relationship, linking the object to its associated functional element. An attribute of this relationship is the affordance label extracted by the 2D detector trained on our generated dataset, which specifies the purpose of the functional element in the context of the object.

\newcommand{\tdepth}{$\theta_{depth}=0.1$ [m]}
\newcommand{\tarea}{$\theta_{depth}=800$ [pixels$^2$]}
\newcommand{\tpoints}{$\theta_{points}=0.6$}
\newcommand{\tbbox}{$\theta_{bbox}=0.4$}
\newcommand{\trarea}{$\theta_{rarea}=0.7$}
\newcommand{\tgeo}{$\theta_{geo}=0.5$}
\newcommand{\tsem}{$\theta_{sem}=0.6$}
\newcommand{\trel}{$\theta_{rel}=1$}
\newcommand{\tgeoo}{$\theta_{geo2}=0.6$}
\newcommand{\tnum}{$\theta_{num}=3$}

\subsection{Implementation details}
For the 2D dataset generation, we use \tdepth, \tarea, and \tpoints~as thresholds.

For the 3D scene graph generation pipeline, we use YOLO-WORLDv8.2 for object detection and RT-DETR for functional element detections, the best in terms of \textbf{mAP}$^{val}_{50:95}$ according to Table~\ref{tab:detection2d}. 
As a general-purpose semantic segmentation model we use SAM2 \cite{sam}, and as VLM GPT-4o \cite{gpt4}. 
The thresholds in this context are \tbbox, \trarea, \trel, \tgeo, \tsem, \tgeoo~and \tnum. All computations are performed on a machine with an NVIDIA 4090 GPU, 64GB RAM + 64GB SWAP, and an AMD Ryzen 9 7950X Processor.

\section{Experiments}\label{sec:experiments}
In the following, we conduct experiments to investigate the accuracy of our trained 2D models and compare them to existing zero-shot detectors. We further validate the scene graph generation and 3D lifting against the current SOTA 3D detectors for functional elements segmentation. Finally, we investigate the usefulness of our generated 3DSGs to respond to task-driven affordance queries such as ``open the left window above the radiator.''

\subsection{Detecting Functional Interactive Elements in Images}\label{sec:exp2ddet}

A fundamental assumption of our proposed pipeline is that functional elements can be detected at least as well from a collection of RGB-D images as from a high-quality laser scan. Therefore, we first compare the 2D detection performance of our trained models and, in the following section, place this into context with existing 3D models. 

Among the nine classes of SceneFun3D, we retained only the seven most appropriate for describing functional elements, discarding ``unplug" and ``plug\_in" along with their respective annotations, as they primarily focused on plugs (we also do not consider them in possible queries and annotations in all the experiments). 
Taking this into account and based on the process described in Section~\ref{sec:methb}, we produce a total of \num{274022} annotations across \num{132635} images with our projection pipeline. We further augmented the data with 1\% background images (images in which no functional element is visible), forming our \textit{standard dataset} (T).
The train-validation split of the dataset is 80/20, with the split ensuring that train and validation images come from different scenes. 

As a baseline, we run YOLO-Worldv8.2~\cite{yolo} and Grounding Dino~\cite{ren2024groundingdino15advance} in a zero-shot fashion. We add the names of the functional elements (``knob," ``handle," ``button," ``switch," ``pedal," and ``general functional interactive element") to their standard prompt set and associate each of their predicted labels with the label of the ground truth bounding box that has the highest overlap with the predicted one, solely to evaluate their detection capabilities.

Next, we fine-tune YOLOv11~\cite{khanam2024yolov11overviewkeyarchitectural} and RT-DETR~\cite{DETRs} on the standard dataset.
 Further, we compare these models on a variant dataset, which we compute using the slicing-aided hyper inference (SAHI) mechanism~\cite{akyon2022sahi}, and refer to it as the \textit{sliced dataset} (ST).
This splits the images into $640\times 640$ patches, which we then use for training and detection. With a slice of 20\% in both directions, where an annotation covers at least 60\% of the original area, this process generates a total of \num{335430} annotated images.

As evident from Table~\ref{tab:detection2d}, zero-shot detections are poor. This is the main reason why, in prior 3DSG works that use such zero-shot models, representing functional elements explicitly is not possible.
According to our evaluation, RT-DETR trained on ST and predicting using SAHI performs best across the metrics. This confirms the findings of~\cite{akyon2022sahi} that SAHI is effective for detecting smaller elements. However, given the little difference between ST and T training data for $\textbf{mAP}_{50}^{\emph{val}}$, we conclude that ST training does not lead to better detections, but to better bounding box accuracy (reflected in \textbf{mAP}$_{50:95}^{\emph{val}}$).

\begin{table}[bt!]
    \footnotesize
    \centering
    \renewcommand{\arraystretch}{1.05}
    \caption{}\vspace{-8pt}
    \setlength{\tabcolsep}{3.85pt}
    \begin{tabular}{l|c|c|c}
        \toprule
        \textbf{Model}  & \textbf{Dataset} & \textbf{mAP}$_{50:95}^{\emph{val}}$ &\textbf{mAP}$_{50}^{\emph{val}}$ \\
        \midrule
        YOLO-Wv8.2 & -  &   0.0   &   0.0     \\
        GroundingDINO    & -  &   0.0   &   0.7    \\
        YOLOv11   & T  &   7.5   &   22.4    \\
        RT-DETR   & T &   8.6   &   \textbf{26.7}    \\
        YOLOv11   & ST &   13.6  &   15.0  \\
        RT-DETR  & ST &   \textbf{21.0}   &   26.2   \\
        \bottomrule
    \end{tabular}
    {\begin{flushleft}
     Fine-tuning results for 2D functional interactive element detection. The dataset column indicates the training dataset: T is \textbf{our} standard training set and ST is \textbf{our} sliced training set. Pre-trained models are used
     if no dataset is listed. We report the standard COCO \cite{coco} metrics.
     \end{flushleft}}
    \label{tab:detection2d}
\end{table}

\subsection{Functional Interactive Elements 3D Segmentation}\label{sec:expfie3dseg}
The next experiment validates the accuracy of the 3D reconstruction and segmentation of the functional element detected in 2D and assesses how much information is lost in the process of transitioning back and forth between 3D and 2D.
To this end, we randomly select 10 scenes from the validation dataset: \texttt{423070}, \texttt{423306}, \texttt{423738}, \texttt{434892}, \texttt{435357}, \texttt{435715}, \texttt{435724}, \texttt{442392}, \texttt{464754}, \texttt{467330}, and associate the point cloud of our functional element node with the eight nearest points within \num{5} [mm] from the original laser scan, for which the ground truth segmentation has been annotated. 
Note that we retain all functional element detections, even without object associations, to avoid penalizing scores when parent objects are undetected.

We then compute the AP metrics as defined in \cite{delitzas2024scenefun3d}. Per-label results are reported in Table~\ref{tab:3dfie}. 
We note that the exact numerical results are difficult to compare, as \cite{delitzas2024scenefun3d} does not release either the model checkpoints or the full train/test split.
Given that the measured performance on the different splits of the same datasets are in a similar range, we carefully conclude that our proposed method achieves similar results to SOTA approaches that directly predict the class for the points in 3D. 
The main penalized classes are those for which it is difficult to generate flawless masks due to background noise. 
Another source of error is not directly related to the method: indeed, the poses $\mathcal{P}$ provided in the dataset \cite{delitzas2024scenefun3d} are not always accurate, generating artifacts in the merging process and penalizing the metrics that require higher precision.

To further emphasize the correlation between 2D detection quality and 3D segmentation performance, we conducted an ablation study to relate the metrics from the previous experiment to the varying prediction confidence threshold $\theta_{bbox}$ of our trained detector. The results are presented in Table~\ref{tab:confidence_performance}.

\begin{table}[bt!]
    \footnotesize
    \centering
    \renewcommand{\arraystretch}{1.05}
    \caption{}\vspace{-8pt}
    \setlength{\tabcolsep}{3.85pt}
    \begin{tabular}{c|c|c|c|c}
        \toprule
        \textbf{Affordance}  & \textbf{AP} & \textbf{AP$_{50}$} & \textbf{AP$_{25}$} & \textbf{AP$_{10}$} \\
        \midrule
        Foot Push   & 0.0  & 0.0  & 20.0  & 46.7 \\
        Tip Push    & 10.0  & 19.9  & 28.4  & 28.4 \\
        Rotate      & 6.1  & 19.9  & 25.6  & 28.6 \\
        Pinch Pull  & 1.7  & 6.6   & 24.8  & 31.5 \\
        Hook Pull   & 0.0  & 0.0   & 10.1  & 10.1 \\
        Hook Turn   & 9.5  & 25.2  & 37.7  & 59.0 \\
        Key Press   & 14.1 & 40.4  & 65.8  & 65.8 \\
        \midrule
        Average & 5.9 & 16.0 & \textbf{30.3} & \textbf{38.6} \\
        Mask3D-F \cite{delitzas2024scenefun3d} & [\textbf{7.9}] & [\textbf{18.3}] & [26.6] & - \\
        \bottomrule
    \end{tabular}
    \\{\begin{flushleft}
Segmentation results of 3D functional elements by affordance label for validation of our 3D lifting approach. The original values of the Mask3D-F model are reported for qualitative comparison; its training/test data, model, and weights are not yet fully public. Note that our approach has to cope with inaccurate poses whereas \cite{delitzas2024scenefun3d} directly segments on the point cloud not affected by these errors.
    \end{flushleft}}
    \label{tab:3dfie}
\end{table}

\begin{table}[bt!]
    \centering
    \footnotesize
    \renewcommand{\arraystretch}{1.1}
    \caption{}\vspace{-8pt}
    \setlength{\tabcolsep}{6pt}
    \begin{tabular}{c|c|c|c|c}
        \toprule
        \textbf{Confidence} & \textbf{AP} & \textbf{AP$_{50}$} & \textbf{AP$_{25}$} & \textbf{AP$_{10}$} \\
        \midrule
        0.2 & 4.4  & 8.8  & 24.2 & 32.3 \\
        0.3 & 5.1  & 10.4 & 27.6 & 34.0 \\
        0.4 & \textbf{5.9}  & \textbf{16.0} & \textbf{30.3} & \textbf{38.6} \\
        0.5 & 5.0  & 10.8 & 27.6 & 37.2 \\
        0.6 & 3.6  & 8.0  & 22.1 & 34.9 \\
        \bottomrule
    \end{tabular}
    \label{tab:confidence_performance}
    \\{\begin{flushleft}
    Ablation study on the impact of 2D object detection confidence for 3D segmentation performance.
    \end{flushleft}}
\end{table}

\subsection{Affordance Grounding}
In the following experiment, we aim to assess the performance of our method on the task-driven affordance grounding benchmark in \cite{delitzas2024scenefun3d}: 
each query consists of a text-based description of the task; the goal is to localize and segment the functional interactive elements that an agent needs to interact with in order to successfully accomplish the task. 
For example, the expected result of ``Open the trash bin" is the point cloud of all the trash can pedals in the scene.

To achieve this, we convert our 3D scene graph representation into a JSON format, retaining information about each node's ID, 3D center of mass, 3D bounding box extension, label, relationships with the environment, and functionality affordance if it is a functional element.

We then instruct GPT-4o to find in the JSON the ID(s) of the node(s) that solve the query. This highlights the general advantage of 3DSG representations as they can be easily parsed by LLMs.
On the same set of scenes, and in the same manner discussed in \ref{sec:expfie3dseg}, we retrieve the closest points to our prediction in the original point cloud and compute the 3D point cloud intersection over union (IoU) between our prediction and the ground truth answer elements. We count a query as passed if the IoU is at least 25\% (AP$_{25}$).

In Table~\ref{tab:taskgrounding}, we report per-scene results and compare them to the SOTA ConceptGraphs \cite{conceptgraph} that can answer unconstrained language queries on the map. 
As is evident from the numbers, ConceptGraphs does not account for the possibility of providing functional elements as answers to queries. Instead, it returns whole object point clouds, which results in low IoU with the ground truth.
Therefore, we further report AP$_{>0}$, where we count a query as successful if there is even a single overlap point between the response and the ground truth.
The results, however, show that the return of ConceptGraphs is still less accurate, indicating that the inclusion of functional elements and object-part relations in the 3DSG not only improves retrieval localization but also generally allows for answering more queries correctly. 

Interestingly, one of the main advantages of storing segmented functional elements in the 3DSG, is that the scores between 3D functional element segmentation and affordance grounding do not differ much because all the information needed is stored and only needs to be identified.
To further contextualize the numbers from Table~\ref{tab:taskgrounding}, the AP$_{25}$ of the best model in~\cite{delitzas2024scenefun3d} is 17.5. Even considering the above-mentioned issues of their unavailable models, the comparison indicates that 3DSGs are better representations for answering task-driven queries than direct retrieval from the point cloud.

\begin{table}[bt!]
    \centering
    \footnotesize
    \renewcommand{\arraystretch}{1.1}
    \caption{}\vspace{-8pt}
    \setlength{\tabcolsep}{6pt}
    \begin{tabular}{c|c|c|c|c|c}
        \toprule
        \textbf{Scene} & \textbf{\#Queries} & \multicolumn{2}{c|}{\textbf{ConceptGraphs}} & \multicolumn{2}{c}{\textbf{FunGraph (ours)}} \\
        \midrule
        & & AP$_{25}$ & AP$_{>0}$ & AP$_{25}$ & AP$_{>0}$ \\
        \midrule
        423070 & 8  & 0.0  & 25.0 & \textbf{50.0}  & \textbf{50.0} \\
        423306 & 3  & 0.0  & 0.0  & \textbf{33.3}  &  \textbf{66.7} \\
        423738 & 21  & 0.0  & 57.1  & \textbf{33.3}  & \textbf{85.7} \\
        434892 & 5  & 0.0  & \textbf{40.0}  & \textbf{40.0}  & \textbf{40.0} \\
        435357 & 10  & 0.0  & 50.0  & \textbf{30.0}  & \textbf{60.0} \\
        435715 & 12  & 0.0  & 8.3  & \textbf{33.3}  & \textbf{75.0} \\
        435724 & 10  & 0.0  & 10.0  & \textbf{10.0}  & \textbf{20.0} \\
        442392 & 8  & 0.0  & 25.0  & \textbf{37.5}  & \textbf{37.5} \\
        464754 & 18  & 0.0  & 22.2 & \textbf{27.8}  & \textbf{44.4} \\
        467330 & 4  & 0.0  & 50.0  & \textbf{100}  & \textbf{100} \\
        \midrule
        Total & 99  & 0.0  & 31.3 & \textbf{33.3}  & \textbf{58.6} \\
        \bottomrule
    \end{tabular}
    \\{\begin{flushleft}
    Results for task-driven affordance grounding on randomly selected scenes and queries taken from \cite{delitzas2024scenefun3d}.
    For each method, the percentage of success (IoU at least $25\%$ and $>0\%$) 
    is shown for the scenes in our validation sample. The original number of queries for each scene is reported.
    \end{flushleft}}
    \label{tab:taskgrounding}
\end{table}

\subsection{Label Refinement Ablation Study}
As described in Section~\ref{sec:methb}, the first detection of functional elements only associates them with their affordance label. To obtain also a semantically meaningful label, we prompt GPT-4o with each group of object and its associated functional elements in the image. We call this approach ``GPT-Context" to highlight that the model is aware of the intra-object relationship. We compared this approach to prompting GPT-4o with only the functional element's bounding box annotation (approach ``GPT-No-Context") and using CLIP features to classify it within a closed vocabulary of functional elements. 
We manually verified 100 detected functional objects, labeling them as 'C' (correct), 'W' (wrong), or 'P' (partially correct), where partially correct cases indicate predictions like a ``cabinet drawer knob" instead of a ``cabinet door knob."  Results are reported in Table~\ref{tab:ablation}. CLIP features alone perform poorly due to the small size of functional elements and their bag-of-words behavior, which generates many incorrect results. The best refinements come from the GPT-Context solution, supporting our design choice of early functional-elements-object grouping.

\begin{table}[bt!]
    \footnotesize
    \centering
    \renewcommand{\arraystretch}{1.05}
    \caption{} \vspace{-8pt}
    \setlength{\tabcolsep}{4pt}
    \begin{tabular}{c|ccc|ccc|ccc}
        \toprule
         & \multicolumn{3}{c|}{\textbf{CLIP}} & \multicolumn{3}{c|}{\textbf{GPT-No-Context}} & \multicolumn{3}{c}{\textbf{GPT-Context}} \\
        \midrule
         & C & P & W & C & P & W & C & P & W \\
        \midrule
        Handle   & 32.5 & 37.5 & 30.0 & 77.5 & 12.5 & 10.0 & \textbf{82.5} & 10.0 & 7.5  \\
        Knob     & 43.3 & 33.3 & 23.3 & 23.3 & 56.7 & 20.0 & \textbf{70.0} & 23.3 & 6.7  \\
        Button   & 10.0 & 15.0 & 75.0 & 55.0 & 20.0 & 25.0 & \textbf{85.0} & 10.0 & 5.0  \\
        Other    & 20.0 & 0.0  & 80.0 & 30.0 & 20.0 & 50.0 & \textbf{70.0} & 30.0 & 0.0  \\
        \midrule
        Total    & 30.0 & 28.0 & 42.0 & 52.0 & 28.0 & 20.0 & \textbf{78.0} & 16.0 & 6.0  \\
        \bottomrule
    \end{tabular}
    \\{\begin{flushleft}
Ablation study on context-based label refinement.
Correct (C), partially correct (P), and wrong (W) predictions \% are reported. In total, 100 objects have been studied: 40 handles, 30 knobs, 20 buttons, and 10 other functional elements (pedals, touchscreens, etc.).
    \end{flushleft}}
    \label{tab:ablation}
\end{table}

\section{Limitations}
Even though our approach shows promising results compared to the state of the art in 3D scene graph research for functional element segmentation and retrieval, there is still room for improvement.

We took advantage of recent developments in resources for 3D functional elements to demonstrate how this information can be incorporated into a 3DSG. However, we emphasize that this is only an initial extension of the structure in terms of sub-object parts. 
For example, our method directly relates the knob to the cabinet but not to the drawer of the cabinet, an intermediate information that may still be valuable to retain.

Further exploration is needed to better relate functional elements that may be shared between multiple objects or mounted between two objects. For example, knobs commonly found between an oven and a stove are among the most frequent sources of naming and association errors, according to our analysis.

\section{Conclusions}
In this work, we presented FunGraph, the first 3D scene graph framework that models intra-object relationships, focusing on functional elements to enable tasks requiring interaction with objects in a scene. 

Through our experiments, we demonstrated that our finer-grained representation achieves performance comparable to state-of-the-art 3D detectors, and we highlight the superiority of the method to direct point cloud affordance grounding. However, our approach is fundamentally rooted in the 2D domain. It does not rely on segmenting a pre-existing high-quality point cloud, which makes it also suitable for robotics applications with affordable RGB-D sensing.
We are able to detect and store information about the functional elements of objects while extending the general 3DSG generation pipeline and preserving the graph's hierarchical property.

In the future, we will augment 3D scene graphs with even more fine-grained representations by introducing intermediate object parts before linking the objects themselves to functional elements. Moreover,  we will integrate all the necessary manipulation information into functional element nodes, enabling robots to perform motions that interact with objects and ultimately providing an end-to-end solution.

\bibliographystyle{IEEEtran} 
\bibliography{main}
\end{document}